\pdfoutput=1

\documentclass[11pt]{article}

\usepackage{acl}

\usepackage{times}
\usepackage{latexsym}

\usepackage{graphicx}
\usepackage{framed}
\usepackage[normalem]{ulem}
\usepackage{booktabs}
\usepackage{array}
\usepackage{xspace}
\usepackage{comment}
\usepackage{booktabs}
\usepackage{enumitem}
\usepackage{multirow}
\usepackage[export]{adjustbox}
\usepackage{mathtools}
\usepackage{booktabs}
\usepackage{todonotes}
\usepackage{makecell}
\usepackage{arydshln}
\usepackage{multirow}
\usepackage{makecell}
\usepackage{color, colortbl}

\usepackage{subcaption}

\usepackage{microtype}

\newcommand{\logoB}[0]{\includegraphics[width=.05\textwidth]{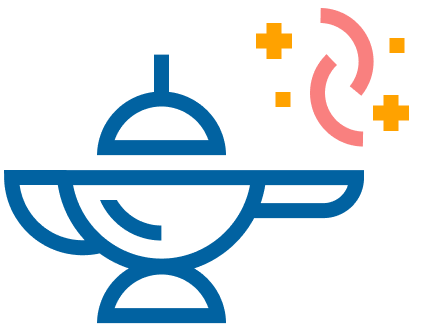}}

\definecolor{purple}{rgb}{0.5,0,1}
\definecolor{dcyan}{rgb}{0.2,0.6,0.5}
\definecolor{light-gray}{gray}{0.95} 

\definecolor{darkgreen}{RGB}{0,140,0}
\definecolor{darkred}{RGB}{200,0,0}
\definecolor{lightgreen}{RGB}{238,247,233}
\definecolor{lightred}{RGB}{255,205,212}
\definecolor{lightyellow}{RGB}{255,240,160}
\definecolor{lightblue}{RGB}{195,221,255}

\newcolumntype{L}[1]{>{\raggedright\let\newline\\\arraybackslash\hspace{0pt}}m{#1}}
\newcolumntype{C}[1]{>{\centering\let\newline\\\arraybackslash\hspace{0pt}}m{#1}}

\makeatletter
\def\adl@drawiv#1#2#3{%
        \hskip.5\tabcolsep
        \xleaders#3{#2.5\@tempdimb #1{1}#2.5\@tempdimb}%
                #2\z@ plus1fil minus1fil\relax
        \hskip.5\tabcolsep}
\newcommand{\cdashlinelr}[1]{%
  \noalign{\vskip\aboverulesep
           \global\let\@dashdrawstore\adl@draw
           \global\let\adl@draw\adl@drawiv}
  \cdashline{#1}
  \noalign{\global\let\adl@draw\@dashdrawstore
           \vskip\belowrulesep}}
\makeatother

\newcommand{\ignore}[1]{}

\newcommand{\changed}[1]{{#1}}

\newcommand{\subscript}[2]{$#1 _ #2$}
\newcommand{\genie}[0]{\textsc{Genie}\xspace}

\definecolor{pinkp}{RGB}{222,108,186}

\usepackage[T1]{fontenc}

\usepackage[utf8]{inputenc}

\usepackage{microtype}

\title{
\vspace*{-0.5in}
{{\small \hfill EMNLP 2022}\\
\vspace*{.25in}} 
\genie\logoB\ Toward Reproducible and Standardized Human Evaluation \\ for Text Generation}

\author{
Daniel Khashabi$^{3}$\thanks{~~Work done at Allen Institute for AI.} $\;$ Gabriel Stanovsky$^{1,4}$ $\;$ \textbf{Jonathan Bragg}$^{1}$ $\;$ \textbf{Nicholas Lourie}$^{5*}$ $\;$ \textbf{Jungo Kasai}$^{2}$  
\\
\textbf{Yejin Choi}$^{1,2}$ $\;$ \textbf{Noah A. Smith}$^{1,2}$ $\;$ \textbf{Daniel S. Weld}$^{1,2}$ $\;$
\\
{
\small
 $^1$Allen Institute for AI \; 
 $^2$University of Washington \; 
 $^3$Johns Hopkins University \; 
 $^4$Hebrew University of Jerusalem \; 
 $^5$New York University
 }
}

\begin{document}
\maketitle
\begin{abstract}
While often assumed a gold standard, effective human evaluation of text generation remains an important, open area for research.
We revisit this problem with a focus on producing consistent evaluations that are \emph{reproducible}---over time and across different populations. 
We study this goal in different stages of the human evaluation pipeline. 
In particular, we consider design choices for the annotation interface used to elicit human judgments and their impact on reproducibility. 
Furthermore, we develop an automated mechanism for maintaining annotator quality via a probabilistic model that detects and excludes noisy annotators. 
Putting these lessons together, we introduce \genie{}: a system for running \emph{standardized} human evaluations across different generation tasks.
We instantiate \genie{} with datasets representing four core challenges in text generation: machine translation, summarization, commonsense reasoning, and machine comprehension.
For each task, \genie{} offers a leaderboard that automatically crowdsources annotations for submissions, evaluating them along axes such as correctness, conciseness, and fluency.
We have made the \genie{} leaderboards publicly available, and have already ranked 50 submissions from 10 different research groups.\footnote{\url{https://genie.apps.allenai.org}} We hope \genie{} encourages further progress toward effective, standardized evaluations for text generation.
\end{abstract}


\section{Introduction}
\label{sec:introduction}

\begin{figure}[tb!]
    \centering
    \includegraphics[scale=0.85,trim=8.5cm 4.2cm 1cm 0.25cm,clip=true]{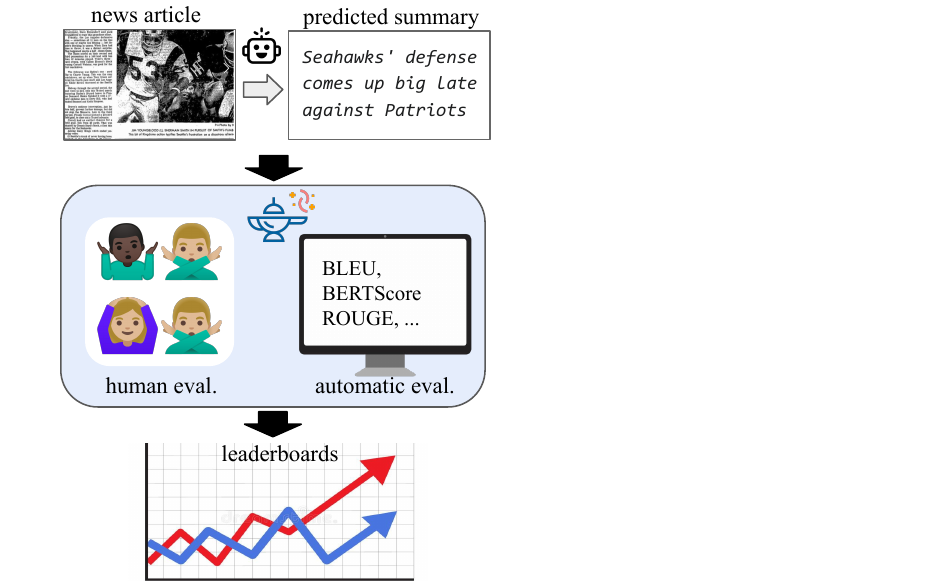}
    \caption{The \genie{} architecture for evaluating text generation tasks, with a summarization example. Similar to automatic leaderboards, model developers submit their predictions (top). \genie{} then evaluates with a standard human evaluation as well as with automatic metrics (center). These scores are then used to rank and track systems' performance across time (bottom).}
    \label{fig:genie:intro}
\end{figure}

While the emergence of powerful language models \cite{radford2019language,raffel2019exploring,lewis2019bart} has made \textit{text generation} omnipresent, effective evaluation of the resulting systems' performance on open-ended generation tasks remains a challenge. This has motivated adoption of human evaluation in recent works~\cite{celikyilmaz2020evaluation,fabbri2021summeval}, even though it poses several challenges~\cite{clark2021all,karpinska2021perils}. First, the estimates of system performance are not \emph{reproducible}---over time and various annotator populations. Additionally, the setups are not \emph{standardized}. Different works use different annotation interfaces, even those working on the same dataset, despite substantial efforts needed for building an appropriate annotation interface and guidelines to extract quality human annotations and filter out noisy annotators.

This work presents an investigation toward reliably repeatable and standardized human evaluation. First and foremost, we study the reproducibility of human annotations, in two stages of the annotation pipeline. We study this goal empirically as a function of various design choices (\S\ref{sec:designing-human-evaluations}) such as the way the judgments are aggregated. We then propose a probabilistic framework for detecting malicious annotators (\S\ref{sec:monitoring-annotation-quality}) and isolating their annotations from the resulting performance estimates.\footnote{Code implementing the model is available at \url{https://github.com/allenai/genie-worker-scoring}.}

Guided by the earlier studies, we present \genie{} (Figure~\ref{fig:genie:intro})---a framework for  human evaluation of text generation, which scales to a variety of tasks and datasets (\S\ref{sec:a-leaderboard-system}). \genie{} posts model predictions to a crowdsourcing platform,\footnote{We use \href{www.mturk.com}{Amazon Mechanical Turk}.} where human annotators evaluate them according to predefined, dataset-specific guidelines. We describe mechanisms introduced into \genie{} to quantify annotator variance and spread the annotations across various days, showing that \genie{} achieves reliable scores on the studied tasks. To show its applicability, we instantiate \genie{} with leaderboards for several popular text generation datasets in English from four diverse tasks---machine translation, question answering, summarization, and commonsense reasoning---and invite developers to extend it with more datasets. Since its deployment, \genie has analyzed and ranked about 50 submissions from 10 different groups across all of our tasks, indicating the interest in standardized human evaluation.

The \genie{} infrastructure opens the door for three avenues of research: (1) \genie{} provides \changed{\emph{developers} of} text-generation models with the ease of the ``leaderboard experience,'' alleviating the evaluation burden while ensuring high-quality, standardized comparison against previous models. (2) \genie{} facilitates the study of \emph{human evaluation interfaces}~\cite{nenkova-passonneau-2004-evaluating,Liu2016EffectiveCA,Bragg2018SproutCT,shapira-etal-2019-crowdsourcing}, addressing challenges such as annotator training,  inter-annotator agreement, and reproducibility, all of which can be integrated into \genie{} to  compare against other evaluation metrics on past and future model submissions. (3) \genie{} helps developers of \emph{automatic evaluation metrics}~\cite{zhang2019bertscore}, by serving as  a hub of model submissions and associated human scores. 
\section{Related Work}
\label{sec:related-work}

We survey relevant work on automatic and human-in-loop evaluation of text generation. See \citet{welty2019metrology,van2019best,celikyilmaz2020evaluation} for further in-depth discussion. 

\vspace{-0.2cm}
\paragraph{(Semi-)automatic Metrics} Many researchers have proposed automated metrics for text generation tasks, such as BLEU~\cite{Papineni2001BleuAM} and METEOR \cite{banerjee2005meteor}. These metrics initially correlated well with human judgments for contemporary models \cite{Papineni2001BleuAM, Doddington2002, Coughlin2003}, though the correspondence breaks down as they become targets for optimization~\cite{callison-burch-etal-2006-evaluating,sun-etal-2019-compare} or as models become increasingly powerful~\cite{ma-etal-2019-results, Edunov2020OnTE}. Several more recent approaches aim to learn automated metrics for text generation tasks, including for image description~\citep{vedantam2015cider}, paraphrasing~\citep{sellam2020bleurt}, and abstractive question answering~\cite{chen-etal-2020-mocha}. Such progress in automatic metrics is incorporated into recent leaderboards \cite{billboard}. We integrate some of these metrics into our proposed system to track their correlation with human evaluation.

\vspace{-0.2cm}
\paragraph{Human Evaluation of Language} Given the limitations of automatic metrics, much prior work has developed ways to conduct human evaluation of language generation in general, and machine translation in particular. Human evaluation for machine translation~\cite{graham-2013-continuous, graham-2014-machine, sakaguchi-val-efficient,freitag2021experts} typically involves crowdsourcing where qualified crowd workers score output translations given the reference text. Results from manual evaluation are used as the primary metric in recent WMT competitions~\cite{wmt2016-findings,wmt2018-findings, wmt-2020-findings}. However, to date, human evaluation efforts are typically conducted (1) on individual tasks such as machine translation, (2)  by individual researchers with potentially varying design decisions, making results incomparable across evaluations, or (3) through shared tasks such as WMT, which force synchronization across teams for evaluation, slowing progress. As a result, most of the research on model development still evaluates models solely on automatic metrics such as BLEU \cite{Papineni2001BleuAM}. \genie\ relaxes these limitations by providing a continually-running leaderboard across language generation tasks with shared high-quality human evaluation templates.

\vspace{-0.1cm}
\paragraph{Human-in-the-loop Evaluation} There are a few recent and concurrent leaderboards that incorporate manual analysis, tending to focus on individual tasks. For example, HYPE~\citep{zhou2019hype} is an evaluation platform for image generation, ChatEval~\citep{sedoc2019chateval} is an evaluation platform for chatbots, and, more recently, \citet{zellers2020evaluating} present a leaderboard for the advice generation task introduced in their work. DynaBench~\cite{kiela-etal-2021-dynabench} is a related multi-task leaderboard but uses changing, adversarially-created datasets that do not support our goal of controlled model comparison across time. HUME~\citep{hashimoto-etal-2019-unifying} was proposed as an evaluation metric for summarization and dialog which combines human annotations with automatic metrics for diversity and quality. STORIUM~\cite{akoury2020storium} was introduced for human-in-the-loop generation and evaluation of long open-ended stories as a computer game. Concurrently, \citet{GEM} introduced GEM, a workshop for participant-driven evaluation of language generation tasks. While such workshops inspire progress toward common goals, synchronized evaluations, often only once per year, likely slow progress. We take the view that evaluations on a more frequent, rolling basis will give researchers more flexibility. To the best of our knowledge, \genie is the first crowdsourced human-in-the-loop system that supports task leaderboards and is backed by principled design to ensure scoring reliability of human evaluations.
\section{\genie Principles for Human Evaluation of Generative Models}
\label{sec:the-genie-principles}

There are many ways to run human evaluations. Reflecting on what's needed to compare text generation models across time, we formulated the following principles to guide our design choices.

\paragraph{Application-Motivated} Ultimately, the evaluation's purpose is to identify useful models and techniques. 
Thus, it should measure something informative about the their usefulness in applications (such as a generated text's correctness or fluency).  

\paragraph{Reproducible} To compare different models over time, the evaluation must be reproducible. If repeated, it should give largely the same results. For example, results should hold across different groups of annotators, and remain stable across appropriate lengths of time.

\paragraph{Interpretable} The evaluation should help a researcher understand how the system behaves, and thus must measure an aspect of the system that is easy to understand. An evaluation which ranks models but isn't interpretable has limited usefulness, since different applications might prioritize different things and researchers must navigate cost-benefit trade offs between more expensive, higher performing models and cheaper ones.

\paragraph{Scalar} 
The evaluation should produce an absolute scalar measurement of the model performance (rather than a relative or comparative one) that facilitates comparison of a new model to all those previously evaluated. 

\paragraph{Quantified Uncertainty}
All measurements are subject to uncertainty, including human evaluations. Thus, when comparing evaluations, we should consider how confident we can be that the resulting measurement is close to the true, latent measurement based on a more complete population of inputs and human annotators.

\paragraph{Rolling} 
Given rapid recent advances in natural language generation, it is essential to develop easily-accessible evaluation platforms for frequent model evaluations that do not require competing teams to synchronize with each other.

\paragraph{Extensible} Evaluation of NLP models is actively evolving, as new datasets are introduced and more is learned about how best to conduct human evaluation. 
Therefore, an evaluation framework should be easily extensible to new tasks or the latest practices. 

Next, we empirically study design decisions along the aforementioned evaluation desiderata.
\section{Design Decisions for Consistent Human Evaluations}
\label{sec:designing-human-evaluations}

\begin{figure*}[ht]
    \centering
    \includegraphics[width=0.99\textwidth,trim=0cm 1.2cm 0cm 1.4cm]{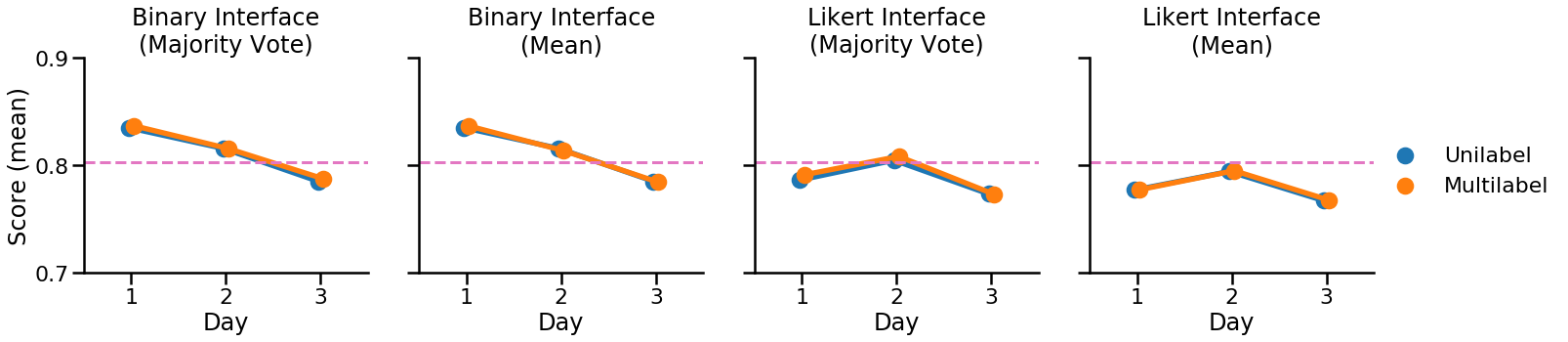}
    \caption{Label aggregation variants across three different days for the \genie{} ARC-DA leaderboard. Using Likert scales yield lower inter-day variation. The horizontal {\color{pinkp}red} dashed line \changed{(0.803) denotes the annotations by an expert annotator intimately familiar with the task}.}
    \label{fig:relabel}
\end{figure*}

\begin{figure}[t]
    \centering
    \includegraphics[width=\columnwidth,trim=0cm 1cm 0.5cm 0.9cm]{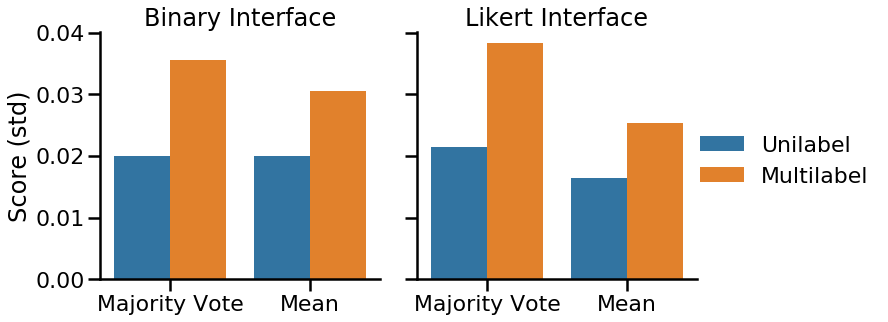}
    \caption{
    \changed{Standard deviation (STD)}
    of different labeling strategies. 
    \emph{Unilabeling} 
    yields lower variance \changed{and hence, 
    better stability across different populations of annotators (on different days).}
    }
    \label{fig:relabel-variance}
\end{figure}

When designing an evaluation, some questions can be answered with principles, while others must be answered empirically. We investigate several questions around the prompt design that commonly occur across various tasks and impact evaluations' reproducibility and confidence.
\begin{enumerate}[nolistsep, label=(\subscript{\emph{\textbf{Q}}}{{\arabic*}})]
    \item \label{q1} \emph{Granularity of the elicitation:} We examine two kinds of labels: (a) binary, and (b) Likert for 5 categories: Strongly agree, agree, neutral, disagree, and strongly disagree.
    \item \label{q2} \emph{Aggregation of per-example labels:} Given multiple labels per example, we investigate aggregating by (a) averaging their scores, and (b) taking a majority vote. 
    \item \label{q3} \emph{Labels per example:} for a fixed annotation budget, we compare collecting (a) 3 labels per annotation example (multilabeling), with (b) one label for three times\footnote{\changed{``3 times'' is to ensure the same amount is annotated in both scenarios for fair comparison.}} as many annotation examples (unilabeling). 
\end{enumerate}

\paragraph{Case Study: Comparing Evaluation Designs \changed{for Open-domain Question Answering} (ARC-DA} To study these design 
choices,
we evaluated T5-11B~\cite{raffel2019exploring} on the development set of ARC-DA~\cite{clark2021all}\changed{, a generative question answering dataset (see \S\ref{subsec:datasets} further details)}. We used modified versions of the same annotation interface as \citet{bhakthavatsalam2021think} (Figure~\ref{fig:templates}). Each evaluation was run once with a Likert scale and once with a binary scale. All instances ($n=360$) were annotated by 3 annotators, repeated three times across different weekdays.\footnote{For consistency, experiments were launched at 10am PST.} Then the quality judgments were mapped to numerical values.\footnote{\label{ft:scores} The binary scale was mapped to 0 and 1, while the Likert scale was mapped to 0, 0.25, 0.50, 0.75, and 1.} To produce the unilabeling and multilabeling results, we simulated these policies by randomly sampling with replacement for 500 rounds, either a random 1/3 of the total number of examples (multilabeling) or 1/3 of the total number of annotations for each example (unilabeling).\footnote{This budget ensured that the number of sampled examples was at most the total number of examples, for unilabeling.}

Figure~\ref{fig:relabel} compares the reproducibility of different setups across time. Each subplot represents a choice of \ref{q1} scale (binary/Likert), and \ref{q2} aggregation (mean/majority-vote). We compare these setups across subplots, and within subplots compare \ref{q3} unilabeling and multilabeling. The choices of scale and aggregation appear to have little effect on the evaluation, with all combinations broadly stable across days, though Likert elicitation with mean aggregation is slightly more stable.

Figure~\ref{fig:relabel-variance} compares the variance for all possible combinations. The choices of scale and aggregation appear to have little effect, though the Likert scale with mean aggregation may have the lowest variance. The biggest impact comes from unilabeling, which noticeably reduces the variance in comparison to multilabeling across all scenarios. This observation is consistent with previous work demonstrating the effectiveness of unilabeling for model training~\cite{lin2014re}, but deviates from how annotations are often done in NLP~\cite{van2019best}. Our finding suggests that unilabeling is a promising strategy for model evaluation.

Overall, \emph{unilabeling} with \emph{Likert scales} and \emph{mean aggregation} appears most reliable among all configurations for ARC-DA, and therefore we use this configuration in \genie{}. Moreover, for the main leaderboard evaluations we use 3--7 times more samples, and expect even less variation. Our analysis shows that these design choices provide a good starting point for reproducible experiments with confident estimates.
\section{Monitoring Annotation Quality}
\label{sec:monitoring-annotation-quality}

Despite strict qualification requirements, in our early experiments some annotators chose arbitrary labels after initially choosing correct ones. While a small percentage, these annotators complete a disproportionate share of tasks and significantly impact evaluations. To solve this problem, we built a monitoring system with two components: automatically generated \textit{test questions} and an unsupervised \textit{scoring model}.\footnote{We also tried clustering approaches that don't require test questions; however, they did not have good performance.}

\paragraph{Test Questions} Because noisy annotators could favor marking examples as correct or incorrect, test questions need both \textit{positive} and \textit{negative} examples. For positive examples, we replaced model predictions with gold responses. For negative examples, we cyclically permuted the gold generations, so no example was matched with its original. Thus, the negative examples look correct at a glance, but almost never are.

\paragraph{Scoring Model} Manually reviewing annotations can be time consuming and error prone, so we automate this process with a \textit{scoring model} to infer if workers have acceptable accuracy. Probabilistic models of annotation have been richly studied \citep{hovy2013learning, passonneau2014benefits, paun2018comparing}. Much prior work uses worker agreement to identify noisy annotators. Since we use unilabeling (\S\ref{sec:designing-human-evaluations}), workers annotate disjoint sets of examples and these methods are not applicable. Instead, we use a similar probabilistic model but applied to predict how often workers correctly answer the test questions. Such a model must be \textit{unsupervised}, since new tasks won't have identified noisy annotators, \textit{interpretable}, since parameters like confidence thresholds must be set a priori, and \textit{sequential}, so noisy annotators can be detected as soon as there is enough evidence.

In our model, each worker, $w$, answers $n_w$ test questions. The number of correctly answered test questions, $X_w$, is binomially distributed with mean $P_w$. Each $P_w$ comes from a mixture of betas prior. Thus, noisy and non-noisy annotators can be modeled with different mixture components.
\begin{align*}
   Z_w &\sim \text{Categorical}(\theta_1, \dots, \theta_k) \\
   P_w &\sim \text{Beta}(\alpha_{Z_w}, \beta_{Z_w}) \\
   X_w &\sim \text{Binomial}(P_w, n_w)
\end{align*}
We compare two definitions of noisy annotators. The \textbf{rate} \changed{criterion} defines them as workers with an accuracy ($P_w$) below a threshold ($90\%$). The \textbf{class} criterion defines them as workers whose latent class ($Z_w$) corresponds to any mixture component besides the one with highest expected accuracy.

We fit the model parameters, $\theta_i, \alpha_i, \beta_i$, for mixture components $i=1, \dots k$, with maximum likelihood via the EM algorithm~\cite{dempster1977maximum} for mixture models~\citep{murphy2012probabilistic}. Then, we infer a posterior distribution for each worker's accuracy ($P_w$) and latent class ($Z_w$) given the number of questions they answered correctly ($X_w$). Since the prior is a mixture of conjugate distributions, the posteriors have a closed-form~\citep{diaconis1979conjugate}.

To adapt the Likert responses for this model, we binarize them at 0.5. Positive and negative test questions are modeled independently, and annotators are considered noisy if they are noisy on either. Since the difficulty of annotating different tasks varies, each \genie{} task is modeled separately. Finally, to stabilize the EM algorithm and resulting parameter estimates, we augment the worker responses with pseudo-data. See Appendix~\ref{app:monitoring-annotation-quality:modeling} for the full technical details.

\paragraph{Detecting Noisy Annotators for WMT21} To test \genie{} in a real-world scenario, we used it to evaluate the 24 systems submitted to WMT21 and several additional baselines on German-to-English translation \citep{akhbardeh-etal-2021-findings}. For the evaluations, the \genie{} leaderboards used 5\% of examples as positive and 5\% as negative test questions. We manually reviewed test question statistics to identify and remove 5 noisy annotators from a pool of 88 (5.7\%). As in our preliminary experiments, these noisy annotators represented a small fraction of annotators; however, we had previously found such annotators could annotate up to 50\% of the HITs.\footnote{\changed{A HIT (Human Intelligence Task) represents a single, self-contained, virtual task that a crowd worker can work on and collect a reward for completing.}} By identifying and removing them, we prevented such a negative impact on our WMT evaluations.

\paragraph{Simulation Study} Even a fairly large real-world evaluation encounters only a few noisy annotators. So, we complement our WMT21 case study with simulations based on it, where we can run more trials and know the ground truth.

We split the WMT21 annotations chronologically into validation and test sets. The validation set was used during model development, while we evaluated the models by simulating 25 rounds of annotation based on the test set's statistics.\footnote{The test set contained only 2 noisy annotators, too few to compute reliable metrics in a direct evaluation.} Similarly to the annotation models discussed in \citet{karger2011Budget}, each worker was independently designated as noisy and then assigned a rate at which they labeled test questions correctly. Based on the validation data's statistics, for each round we drew the noisy annotator probability uniformly from 1--10\%, and each annotator's probability of being correct uniformly between 0--50\% for noisy annotators and 95--100\% for the rest. The model predicted annotators as noisy if the posteriors for $Z_w$ and $P_w$ assigned them at least 99\% probability of being noisy annotators under the \textbf{rate} or \textbf{class} criteria. We computed precision and recall across all the simulations, bucketing workers by how many test questions they answered.

\begin{table}[ht]
    \small
    \centering
    \resizebox{0.96\linewidth}{!}{
    \begin{tabular}{lllccc}
        \toprule
        \textbf{Model}         & \textbf{Prior} & \textbf{k} & \multicolumn{3}{c}{\textbf{Precision} / \textbf{Recall}} \\
                               &                &            &      \textbf{1-4} &     \textbf{5-14} &     \textbf{15+} \\
        \midrule
        \multirow{2}{*}{class} & fixed     & 2 & 100/~24 & 100/~91 & 100/~87 \\
                               & learned   & 2 & 100/~15 & 100/~77 & 100/100 \\
        \hline
        \multirow{6}{*}{rate}  & Jeffreys  & 1 & ~68/~85 & ~93/~93 & 100/~98 \\
                               & uniform   & 1 & ~56/~86 & ~93/100 & 100/100 \\
                               & fixed     & 1 & 100/~32 & 100/~94 & ~98/100 \\
                               & learned   & 1 & 100/~14 & 100/100 & ~96/100 \\
                               & fixed     & 2 & ~86/~33 & 100/100 & 100/~97 \\
                               & learned   & 2 & 100/~12 & 100/~92 & 100/100 \\
        \bottomrule
    \end{tabular}
    }
    \caption{Noisy annotator detection models' precision and recall for workers who answered different numbers of test questions (1-4, 5-14, and 15+). Precision and recall were averaged across multiple simulations.
    }
    \label{tab:simulation-results}
\end{table}

Table~\ref{tab:simulation-results} shows the simulation results. In addition to varying the number of components (\textbf{k}) in the \textit{learned} priors, we also compared against uninformative priors (the \textit{Jeffreys} and \textit{uniform} priors), and informative priors (\textit{fixed}). Noisy annotators lose the chance to answer additional test questions, thus it's critical that models have high \textit{precision} when marking workers as noisy. The uninformative priors suffer from low precision, assigning too much probability to a worker being a noisy annotator. The informed and learned priors both perform well, with high precision and good recall---in some cases identifying almost all noisy annotators with fewer than 15 test questions. The learned priors have the additional advantage that they can adapt to different distributions by pooling information across annotators. Based on these results, the 2-component learned rate and class models have proven to be strong candidates for application.
\section{Automatically Managing Human Evaluation Leaderboards}
\label{sec:a-leaderboard-system}

This section reviews the \genie{} system, which automates much of the management of text generation leaderboards with human evaluations. While we note that some human management, such as providing support and handling disputes, should never be fully automated, \genie\ alleviates much of the overall burden.  The next section (\S\ref{sec:the-genie-leaderboards}) describes its instantiation into the \genie{} leaderboards for four diverse text generation tasks.

At a high level, the \genie{} system coordinates a leaderboard UI, data processing backend, and crowdsourcing campaigns on Amazon Mechanical Turk. After retrieving newly uploaded submissions, the backend computes automatic metrics. Upon success, the backend then creates annotation tasks on Amazon Mechanical Turk (AMT) using AMTI\footnote{\url{https://github.com/allenai/amti}} (A Mechanical Turk Inferface), an open-source Python package for working with AMT. 

Each leaderboard is a separate instance of the system, with its own crowdsourcing templates, including instructions, examples, and prompts (see \S\ref{sec:the-genie-leaderboards}). Following our observations in \S\ref{sec:designing-human-evaluations}, all templates use Likert scales which are then mapped to real-valued scores (cf.~footnote~\ref{ft:scores}) and averaged.

The system also maintains a history of past annotations (per-instance and per-worker), updating statistics after each evaluation. This has several immediate and future benefits: worker statistics enable spam detection (\S\ref{sec:monitoring-annotation-quality}), while the annotations can be used for future studies on human evaluation.

These components enable the following features:

\paragraph{Extensibility} New tasks can be modularly added to the \genie{} system, creating new leaderboards. Each task requires a crowdsourcing template and a code object specifying how to push model predictions into and pull workers' annotations from the crowdsourcing templates.
We release an extensible open-source annotation template library,\footnote{\url{https://github.com/allenai/evaluation-interfaces}} seeded with the four task templates used in this work.

\paragraph{Uncertainty Quantification} To better inform model comparisons, we report scores with uncertainty estimates. Bootstrap resampling (samples with replacement from the observed annotations) provides the 95\% confidence intervals for the estimated submission quality scores, as commonly done in machine translation \cite{koehn-2004-statistical}.

\paragraph{Human Evaluations: Uncertainty vs Cost} 
To balance confidence with affordability, the system evaluates a subsample of the test sets. This subset is random, but fixed to reduce the variance between model comparisons. Sentence-level tasks, such as translation of sentences, cost less to annotate per example.
\changed{Depending on task difficulty, we adjust the pay rate per HIT such that we are paying workers at a higher rate than 15 USD per hour.}
For these tasks we annotate 800 instances at a cost of $\sim$\$600 per submission (standard error  $<1.77$\%). For larger tasks, we evaluate 300 instances costing $\sim$\$350 per submission (standard error $<2.89$\%).\footnote{See Appendix~\ref{app:standard-error} for a discussion of standard error.} These evaluations are much larger than what was previously done, e.g., 100 instances for MT in \citet{ma2018results} or around 100 instances for summarization \cite{kim2019abstractive, hardy2019highres, kryscinski-etal-2019-neural, fabbri2021summeval}.

\paragraph{Automatic Metrics} To supplement human evaluations, we compute recent and popular automatic metrics for each task: METEOR~\cite{banerjee2005meteor}, ROUGE~\cite{lin2006information},  BLEU~\cite{Papineni2001BleuAM}, SacreBLEU~\cite{post2018call}, BLEURT~\cite{sellam2020bleurt} and BERTScore \cite{zhang2019bertscore}. Integrating these metrics into \genie{} enables researchers to examine their correlation with human judgments as well as observing trends as more models are submitted.

\paragraph{Quality Control} To ensure annotation quality, annotators must pass strict qualifications requirements\footnote{I.e., 5000 completed HITs, a 99\% assignment approval rate, and being based in a country with a population predomninantly of native English speakers (e.g., USA, Canada, UK, Australia) since our initial set of tasks focuses on English.} and task-specific qualification tests based on a subset of the questions derived from the task's training data. These tests check that the workers have carefully read the instructions and are comfortable with annotating instances of the particular task. In addition, we replace 5\% of examples with positive and another 5\% with negative test questions, which we analyze with the 2-component learned class model between submission evaluations, as described in \S\ref{sec:monitoring-annotation-quality}. Accordingly, noisy annotations are excluded from results and annotators from the pool of eligible workers. Lastly, to eliminate variability from evaluating at different times (weekend vs.\ weekdays, different work hours), we publish the  AMT tasks on weekdays at 10am Pacific Time.
\begin{figure*}[tb!]
    \centering
     \vspace{-0.1cm}
     \begin{subfigure}[b]{0.97\textwidth}
         \centering
         \caption{Question Answering (ARC-DA). }
         \label{fig:template_arcda}
         \vspace{-0.1cm}
         \fbox{\includegraphics[width=\linewidth]{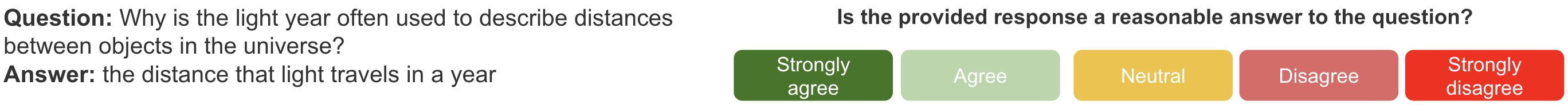}}
         \vspace{0.001cm}
     \end{subfigure}
     \begin{subfigure}[b]{0.97\textwidth}
         \centering
         \caption{Commonsense Reasoning ($\alpha$NLG), adapted from~\citet{Bhagavatula2020Abductive}. }
         \label{fig:template_anlg}
         \vspace{-0.1cm}
         \fbox{\includegraphics[width=\linewidth]{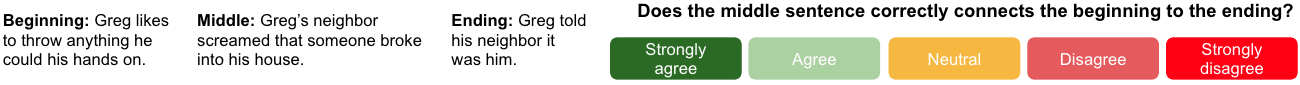}}
         \vspace{0.001cm}
     \end{subfigure}
     \begin{subfigure}[b]{0.97\textwidth}
         \centering
         \caption{Machine Translation (WMT19 and WMT21 DE-EN), adapted from~\citet{wmt2019-findings}. }
         \label{fig:template_wmt}
         \vspace{-0.1cm}
         \fbox{\includegraphics[width=\linewidth]{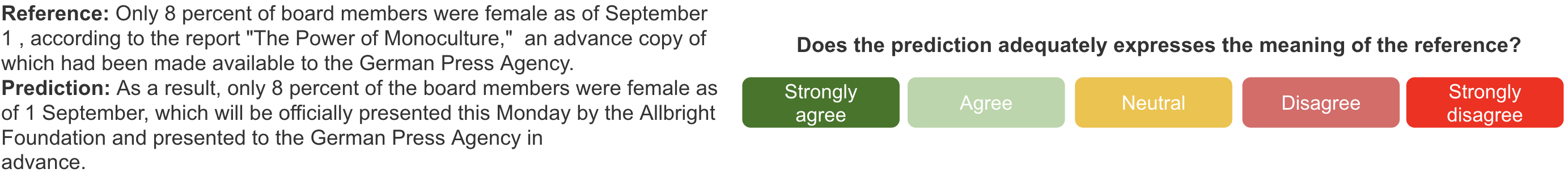}}
         \vspace{0.001cm}
     \end{subfigure}
    \begin{subfigure}[b]{0.97\textwidth}
         \centering
         \caption{Summarization (XSUM), adapted from~\citet{chaganty2018price}. Here, Summary A is the gold label while Summary B is model-predicted text. We permute this randomly between instances so that the annotators are blind to which one is gold.}
         \label{fig:template_xsum}
         \vspace{-0.1cm}
         \fbox{\includegraphics[width=\textwidth]{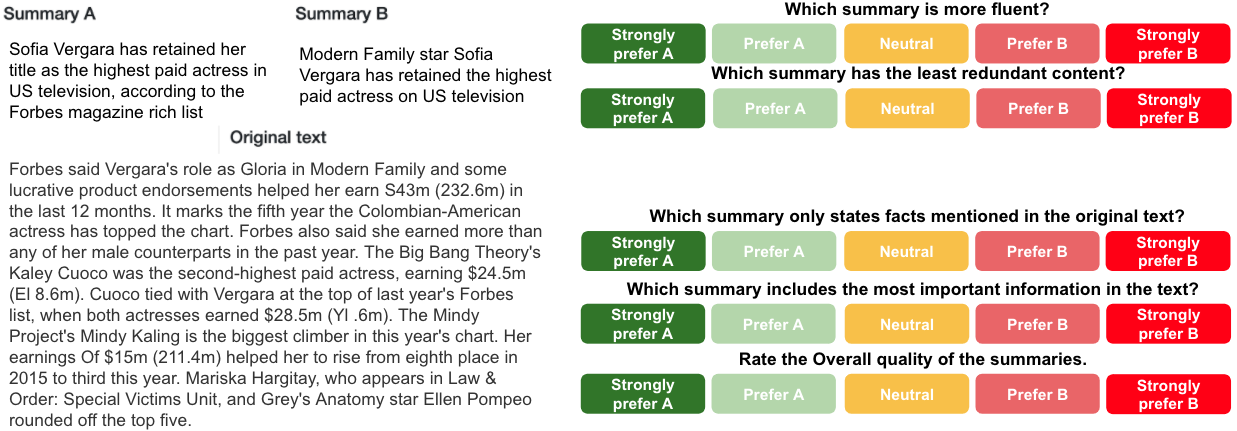}}
     \end{subfigure}
    \caption{Annotation interfaces for the datasets of four tasks integrated in \genie{}. 
    }
    \label{fig:templates}
\end{figure*}


\begin{table}[t]
    \small
    \centering
    \resizebox{0.96\linewidth}{!}{
    \begin{tabular}{@{}cccccc@{}}
        \toprule
        Task           & Dataset                   & Domain & Train & Dev & Test \\ \midrule
        \makecell{Question\\Answering}             & ARC-DA  & \makecell{basic\\science} & 1.4$k$    &  0.4$k$ &  1.5$k$    \\ \cdashlinelr{1-6}
        Summarization  & XSUM                      & News   &  200$k$   & 11$k$  & 11$k$ \\ \cdashlinelr{1-6}
        Commonsense   & aNLG                      & ROCStories  &  170$k$  & 1.5$k$ & 3$k$ \\ \cdashlinelr{1-6}
        \makecell{Machine\\Translation}             & \makecell{WMT19\\DE-EN}                          &   News     &   38.7$m$  &  3$k$     &  3$k$   \\ \cdashlinelr{1-6}
        \makecell{Machine\\Translation}             & \makecell{WMT21\\DE-EN}                          &   News     &   101$m$  &  3$k$     &  1$k$   \\ 
        \bottomrule
    \end{tabular}
    }
    \caption{Datasets currently available in \genie{}, along with their domain and size by task type.}
    \label{tab:datasets-properties}
\end{table}

\section{The \genie{} Leaderboards}
\label{sec:the-genie-leaderboards}

\newcommand{\bl}[1]{{\color{blue}  #1 } }
\newcommand{\tinyf}[1]{{\scriptsize#1}}

\setlength{\tabcolsep}{1.8pt}
\renewcommand{\arraystretch}{0.95}

\definecolor{Gray}{gray}{0.9}
\definecolor{LightCyan}{rgb}{0.88,1,1}

\subsection{Tasks and Datasets}
\label{subsec:datasets}

We integrate in \genie{} datasets from four diverse text-generation tasks, representing longstanding challenges, as outlined below. We focus on  English language datasets, mostly due to easy integration with crowdsourcing platforms. In the future, we hope to integrate other new datasets, particularly other languages. \genie\ is easily extensible; it uses community datasets and metrics via the open-source Datasets library.\footnote{\href{https://github.com/huggingface/datasets}{Huggingface's Datasets repository}} The templates for all tasks are exemplified in Figure~\ref{fig:templates}.

\paragraph{Question Answering} Given an input question about a given context, the system is expected to provide the answer in natural-language form. We use the ARC-DA dataset,\footnote{\href{https://allenai.org/data/arc-da}{ARC-DA dataset}.} which contains questions about subjects from elementary-school science exams. See Figure~\ref{fig:templates} for an example.

\paragraph{Commonsense Reasoning} Given an input scenario, the task is to generate a plausible explanation, according to typical real-world human behavior and understanding. We use $\alpha$NLG~\cite{Bhagavatula2020Abductive}, a dataset for the conditional generation task of explaining given observations in natural language. For evaluation, we use a template and instructions that are similar to those used by \citet{Bhagavatula2020Abductive}, as shown in Figure~\ref{fig:template_anlg}.
    
\paragraph{Machine Translation} The task is to generate a translation in a target language given a text in a source language. Here we use the recent WMT19 and WMT21 datasets with publicly available system outputs \cite{wmt2019-findings, akhbardeh-etal-2021-findings}.\footnote{\label{wmt-preds} \href{https://github.com/wmt-conference/wmt21-news-systems}{WMT21 predictions repository.}} To ensure the generated text is evaluated by native speakers, we focus on German-to-English translation (DE-EN), and leave the expansion to other language pairs as future work. Importantly, WMT19 and WMT21 DE-EN test data only contain text that was originally in German \cite{wmt2019-findings}, avoiding overestimating the quality of translation systems due to translationese effects \cite{toral-etal-2018-attaining, Graham2019TranslationeseIM, Edunov2020OnTE}. We follow the WMT human evaluation template to assess sentence-level translation quality against the reference~\cite{wmt2019-findings}. The one difference is that, consistent with the other \genie{} tasks, we use a five-category Likert scale instead of a continuous one in WMT. See Figure~\ref{fig:template_wmt}.
    
\vspace{-0.2cm}
\paragraph{Summarization} The model is expected to generate a summary of the key points mentioned in a given paragraph. Here we use XSUM~\cite{narayan2018don}, a news summarization dataset. We chose XSUM over alternative datasets for text summarization (e.g., CNN/DM,~\citealp{hermann2015teaching}) since the task involves more abstractive summaries and hence more difficult to evaluate with existing automatic metrics. For evaluating this task we use a template similar to that of \citet{chaganty2018price,fabbri2021summeval} and measure different aspects of quality (redundancy, fluency, conciseness, etc.) that have traditionally been of interest~\cite{mckeown1995generating}. See Figure~\ref{fig:template_xsum} for an example.

\begin{table}[ht!]
    \centering
    \small
    \resizebox{0.96\linewidth}{!}{
    \begin{tabular}{cccccc}
        \toprule
        \multicolumn{5}{c}{ \emph{ARC-DA (Question Answering)} } \\
        \arrayrulecolor{black!30}\midrule
        Systems & Human  & ROUGE & SacreBLEU & BLEURT \\
        \midrule
        \begin{tabular}{@{}c@{}}UnifiedQA \\\tinyf{(ARC-DA/MC+IR)}\end{tabular} & \bl{$\mathbf{80.8}^{+2.1}_{-2.2}$} &  \bf{63.1}	& \bf{22.2}	& \bf{29.40} \\    
        \begin{tabular}{@{}c@{}}UnifiedQA \\\tinyf{(ARC-DA+IR)}\end{tabular} & \bl{${75.3}^{+2.3}_{-2.4}$} &  61.3 & 19.7	& 27.53 \\  
        T5 (11B) & \bl{$66.0^{+2.6}_{-2.5}$}	&  47.4	& 12.8	& 1.6 \\  
        T5 (3B) & \bl{$60.9^{+2.9}_{-3.0}$}		& 43.2  & 11.7 & -5.2  \\
        \arrayrulecolor{black}\toprule
        \multicolumn{5}{c}{ \emph{WMT21 (Machine Translation)} } \\
        \arrayrulecolor{black!30}\midrule
        Systems & Human  & ROUGE  & SacreBLEU & BLEURT \\
        \midrule
        Watermelon & \bl{$\mathbf{75.7}_{-2.0}^{+2.0}$}  & \bf 64.8  & \bf 34.5 & \bf 34.7 \\
        VolcTrans-AT  &  \bl{$\mathbf{75.2}^{+2.0}_{-2.0}$} 		&  \bf 64.8	& 34.4	& 34.6 \\
        HUMAN & \bl{$\mathbf{75.2}^{+2.0}_{-2.0}$}   & 59.3 &29.5&30.0\\
         \genie-large-6-6 & \bl{$70.4^{+1.9}_{-2.0}$}		& 63.3		& 32.4	& 31.3\\ 
         \genie-base-6-6 & \bl{$69.0^{+2.1}_{-2.1}$} 	& 63.3	& 31.8 & 28.2 \\ 
        \genie-base-3-3 & \bl{$65.3^{+2.3}_{-2.3}$}	 & 62.7 & 31.2	& 23.9 \\ 
        \genie-base-1-1 & \bl{$50.7^{+2.3}_{-2.4}$}		& 59.3	& 27.0	& -0.2 \\
        \arrayrulecolor{black}\toprule
        \multicolumn{5}{c}{ \emph{$\alpha$NLG (Commonsense Reasoning)} } \\
        \arrayrulecolor{black!30}\midrule
        Systems & Human  & ROUGE  & SacreBLEU & BLEURT \\
        \midrule
        T5 (11B) & \bl{$\mathbf{{75.9}}^{+1.1}_{-1.0}$} & \bf 44.6  & \bf 19.5 & \bf -22.2 \\
        \begin{tabular}{@{}c@{}}GPT-2 \\\tinyf{(unsupervised)}\end{tabular} & 
        \bl{$45.1^{+1.2}_{-1.3}$} & 19.7  & 1.8 & -84.5 \\

        \arrayrulecolor{black}\toprule
        \multicolumn{5}{c}{ \emph{XSUM (Summarization)} } \\
        \arrayrulecolor{black!30}\midrule
        Systems & 
        \begin{tabular}{@{}c@{}}Human \\overall\end{tabular}  & ROUGE & SacreBLEU & BLEURT \\
        \midrule
         \begin{tabular}{@{}c@{}}Anonymous \\\tinyf{(ARR submission)}\end{tabular} & \bl{$\mathbf{50.6}^{+3}_{-3}$} & 35.8 & 13.9 & -21.7 \\
         Pegasus & \bl{$\mathbf{48.7}^{+3.1}_{-3.4}$} & \bf 39.1  & 16.7 & -17 \\
         T5 (11B) & \bl{$\mathbf{47.5}^{+3.3}_{-3.3}$} & 37.9  & \bf  17.1 & \bf -14.3 \\ 
        \arrayrulecolor{black}\bottomrule
    \end{tabular}
    }
    \caption{Summary of evaluating several existing, strong models on each dataset with \genie{}. The highest numbers and their \changed{confidence interval (CI)}  in each column are indicated in {\bf bold}. The scores given by crowd workers are indicated with {\color{blue}blue} color. We evaluated all participating systems from WMT21 but only show the top 3 systems as well as our \genie transformer baselines for clarity. See Table~\ref{tab:main:experiments:appendix} (appendix) for more metrics and WMT19 results.}
    \label{tab:main:experiments}
\end{table}

\subsection{Evaluating \genie{} Baselines}
\label{sec:experiments}
Here we evaluate several baseline models for each dataset using the \genie{} evaluation pipeline. 

\vspace{-0.2cm}
\paragraph{Models} We use models that are known to perform strongly for each of our tasks.
For all tasks but machine translation, we train and evaluate T5 (11B;~\citealp{raffel2019exploring}), a powerful text-generation model that has shown promising results on a wide variety of text generation tasks. 

For WMT we evaluate other specialized models instead of T5, which is pre-trained only on  English~\cite{raffel2019exploring}. For WMT21 DE-EN, we evaluate all publicly available shared task submissions (see footnote~\ref{wmt-preds}). Additionally, we train and evaluate four transformer-based baselines with varying sizes: \genie{}-large-6-6 (transformer large with a 6-layer encoder and a 6-layer decoder), \genie{}-base-6-6, \genie{}-base-3-3, and \genie{}-base-1-1.\footnote{
\url{https://github.com/jungokasai/GENIE_wmt2021-de-en}
} These models are trained solely on the given training data without ensembling, backtranslation, or any other data augmentation method, to support future research in low-compute settings.

In addition to the above baselines we evaluate specialized baselines for each task. For $\alpha$NLG, we evaluate an unsupervised baseline~\cite{qin2020back} based on GPT-2~\cite{radford2019language}. For summarization, we evaluate Pegasus~\cite{zhang2020pegasus}. For ARC-DA, we evaluate a fine-tuned version of UnifiedQA~\cite{khashabi2020unifiedqa}.

\vspace{-0.1cm}
\paragraph{Results} The results are summarized in Table~\ref{tab:main:experiments}. The human judgment scores for each task are calculated with our described pipeline (\S\ref{sec:a-leaderboard-system}). Even though we have evaluated strong baselines for each task, the machine responses are far from what human judges consider perfect. In the WMT21 task, the transformer baselines are ranked in the expected order: large-6-6, base-6-6, base-3-3, followed by base-1-1. These results support the validity of our evaluations. We defer any further study of the correlations between human judgments and automatic metrics for future work since such a study would require more models to be evaluated. 
\changed{
\section{Limitations}
As with other works which deal with human annotation, the results generated via our evaluation framework will have inherent variability. While we tried to mitigate sources of variation in various ways (see \S\ref{sec:monitoring-annotation-quality},\ref{sec:a-leaderboard-system}), some are bound to remain and are hard to account for. These include, for example, selection bias in the pool of annotators that choose to work on our tasks, who may come from specific countries and social status 
and select for certain tasks and their templates. We welcome future evolution of all parts of the \genie{} architecture, including its evaluation metrics.
}

\section{Conclusion and Future Work}
\label{sec:conclusion}

We introduce \genie{}, a  unified  approach to human-in-the-loop evaluation of text generation over a wide set of text generation tasks. \genie{} is open for use and will be adapted based on future adoption. We encourage submissions from all researchers interested in text generation models. 


\section*{Acknowledgments}
The authors would like to thank the leaderboard team at Allen Institute for AI, particularly Michal Guerquin and Sam Skjonsberg. We thank Peter Clark, Oyvind Tafjord and Daniel Deutsch for valuable feedback throughout this project. 
We are grateful to the many
AMT workers whose contributions make human evaluation possible,
and to the anonymous reviewers for their helpful feedback on this manuscript.
This work was supported in part by DARPA MCS program through NIWC Pacific (N66001-19-2-4031) and research grant 2336 from the Israeli Ministry of Science and Technology.


\bibliography{custom}


\appendix

\clearpage

\section{Details on Model Engineering}
\label{app:models}

Here we summarize the experimental details for building the models used in \S\ref{sec:experiments}. 

\paragraph{T5 models.} For various datasets (except WMT which requires a multi-lingual model) we trained T5 models of different sizes: 11 billion parameters (11B) and 3 billion parameters (3B). We used the default hyperparameters on these frameworks: token-limits of size 512 and 100 for inputs and outputs sequences, respectively; learning rates of 1e-3 and batch sizes of 8. The models  were trained for $100k$ steps on v3-8 TPUs which took about 24 hours to finish, on average. The checkpoint  with the highest score on the dev set of each task was selected for evaluation.

\paragraph{\genie WMT models.} Tables \ref{tab:base-setting} and \ref{tab:large-setting} list hyper-parameters for our \genie transformer baselines. BPE with 32K operations is applied jointly to German and English text. All embeddings are shared.
\begin{table}[ht]
    \small
    \centering
    \begin{tabular}{@{} l@{\hspace{-0.2cm}} r @{}}
        \toprule[.1em]
        \textbf{Hyperparameter} & \textbf{Value}\\
        \midrule[.1em]
        label smoothing & 0.1\\
        \# max tokens & 1024 \\
        dropout rate & 0.1 \\
        encoder embedding dim  & 512\\
        encoder ffn dim  & 2048\\
        \# encoder attn heads & 8\\
        decoder embedding dim  & 512\\
        decoder ffn dim  & 2048\\
        \# decoder attn heads & 8\\
        max source positions & 1024 \\
        max target positions & 1024 \\
        Adam lrate& $5\times 10^{-4}$ \\
        Adam $\beta_1$& 0.9\\
        Adam $\beta_2$& 0.98\\
        lr-scheduler &  inverse square \\
        warm-up lr & $1\times 10^{-7}$ \\
        \# warmup updates & 4000 \\
        max epoch &  7 \\
        \# GPUs &  8 \\
        length penalty & 0.6\\
        \bottomrule[.1em]
    \end{tabular}
    \caption{Transformer-base \texttt{fairseq} hyperparameters and setting.}
    \label{tab:base-setting}
\end{table}

\begin{table}[ht]
    \small
    \centering
    \begin{tabular}{@{} l@{\hspace{-0.2cm}} r @{}}
        \toprule[.1em]
        \textbf{Hyperparameter} & \textbf{Value}\\
        \midrule[.1em]
        label smoothing & 0.1\\
        \# max tokens & 4096 \\
        dropout rate & 0.1\\
        encoder embedding dim  & 1024\\
        encoder ffn dim  & 4096\\
        \# encoder attn heads & 16\\
        decoder embedding dim  & 1024\\
        decoder ffn dim  & 4096\\
        \# decoder attn heads & 16\\
        max source positions & 1024 \\
        max target positions & 1024 \\
        Adam lrate& $5\times 10^{-4}$ \\
        Adam $\beta_1$& 0.9\\
        Adam $\beta_2$& 0.98\\
        lr-scheduler &  inverse square \\
        warm-up lr & $1\times 10^{-7}$ \\
        \# warmup updates & 4000 \\
        max epoch &  7 \\
        \# GPUs &  8 \\
        length penalty & 0.6\\
        \bottomrule[.1em]
    \end{tabular}
    \caption{Transformer-large \texttt{fairseq} hyperprameters.}
    \label{tab:large-setting}
\end{table}
\section{Monitoring Annotation Quality}
\label{app:monitoring-annotation-quality}

This appendix provides details on model construction and evaluation for \S\ref{sec:monitoring-annotation-quality}.

\subsection{Modeling}
\label{app:monitoring-annotation-quality:modeling}

All \textit{rate} models used $P_w = 0.9$ as the threshold for defining a noisy annotator. All \textit{class} models used the mixture component with highest average accuracy to define non-noisy annotators. Workers were marked as noisy if the model assigned more than $99\%$ probability to them being so. For reproducibility, we set Python and NumPy's random seeds to 0 at the beginning of our experiments.

\paragraph{Uninformative Priors} Both uninformative priors had one beta mixture component. The \textit{Jeffreys} model used a Jeffreys prior, or $\text{Beta}(\frac{1}{2}, \frac{1}{2})$. The \textit{uniform} model used a uniform prior, or $\text{Beta}(1, 1)$.

\paragraph{Informative Priors} The 1-component fixed rate model had one beta mixture component with parameters $\alpha = 4$ and $\beta = 1$. Both the 2-component fixed rate model and the 2-component fixed class model had two mixture components with probabilities $0.05$ and $0.95$ and parameters $\alpha = 0.5$, $\beta = 4.5$ and $\alpha = 9.5$, $\beta = 0.5$.

\paragraph{Learned Priors} The learned prior models were all fit via the EM algorithm, as described below, and we tried 1 and 2 components for the rate model and 2 components for the class model.

\paragraph{Optimization} To stabilize the EM algorithm and regularize the parameter estimates, we augmented with pseudo-data. To the data, we added 40 pseudo-workers, each completing 20 tasks: 36 annotators with 19 successes, and four noisy annotators with 1, 1, 5, and 10 successes, respectively. The EM algorithm was run with 10 initializations, each with up to 1,000 iterations and relative tolerance of $1e{-}6$ for stopping. Components were initialized with equal mixture probabilities, uniformly random means from 0 to 1, and concentration parameters drawn from a gamma distribution with a shape parameter of 2. Beta mixture components were fit using the Dirichlet-multinomial fixed point iterator from \citet{minka00estimatinga}, 10,000 iterations and a relative tolerance of $1e{-}7$.

\subsection{Evaluation}
\label{app:monitoring-annotation-quality:evaluation}

For evaluation, we simulated 25 rounds of annotation. In each round, the number of test questions were the counts from the test set of annotations, a fixed noisy annotator rate was drawn uniformly from $1\%$ to $10\%$, a mean and concentration parameter for the beta distribution of noisy annotator's success probabilities was respectively drawn uniformly from $0$ to $0.5$ and $5$ to $50$, and a mean and concentration parameter for the beta distribution of regular annotator's success probabilities was respectively drawn uniformly from $0.95$ to $1$ and $100$ to $1,000$. Each worker was assigned a noisy or regular annotator label and accordingly a success probability, then successes and failures were binomial distributed.
\section{Standard Error}
\label{app:standard-error}

\textit{Standard error} quantifies the variability of an estimate, $\hat{\theta}$. Mathematically, the standard error is the estimate's standard deviation (as opposed to the standard deviation of a single sample). Often, the estimate is an average of multiple, independent samples, in which case the standard error is:
\[ \frac{\sigma}{\sqrt{n}} \]
where $n$ is the number of samples and $\sigma$ is the standard deviation of a single sample. When the estimate is an average, it's approximately normally distributed due to the central limit theorem, making $\hat{\theta} \pm 1.96 \frac{\sigma}{\sqrt{n}}$ an approximate 95\% confidence interval.

\textit{The Bhatia-Davis inequality} \citep{BhatiaDavis} bounds the variance of a random variable in terms of its upper bound, $M$, lower bound, $m$, and expectation, $\mu$:
\[ \sigma^2 \leq (M - \mu) (\mu - m) \]
Since the scores from our annotators are bounded between 0 and 1, the maximum standard deviation for any of them is $0.5$. Moreover, if a model's score is $0.8$ on average, then the maximum standard deviation for its annotations is $\sqrt{(1 - 0.8)(0.8 - 0)} = 0.4$. Dividing by $\sqrt{n}$ translates these into bounds on the worst-case standard error of our estimates:
\[ \operatorname{Standard Error} \leq \sqrt{\frac{\mu(1 - \mu)}{n}},\]
where $\mu$ is the expected score.


\begin{table*}[ht]
    \centering
    \small
    \begin{tabular}{ccccccc}
        \toprule
        \multicolumn{7}{c}{ \emph{ARC-DA (Question Answering)} } \\
        \midrule
        Systems & Human & BERTScore & ROUGE & METEOR & SacreBLEU & BLEURT \\
        \midrule
        T5 (11B) & \bl{$\mathbf{66.0}^{+2.6}_{-2.5}$} & \bf{92.4}	&  \bf{47.4}	& \bf{33.1}	& \bf{12.8}	& \bf{1.6} \\  
        T5 (3B) & \bl{$60.9^{+2.9}_{-3.0}$}	& 91.9	& 43.2 & 30.3 & 11.7 & -5.2  \\
    \end{tabular}
    \begin{tabular}{ccccccc}
        \toprule
        \multicolumn{7}{c}{ \emph{WMT19 (Machine Translation)} } \\
        \midrule
        Systems & Human & BERTScore & ROUGE & METEOR & SacreBLEU & BLEURT \\
        \midrule
        FAIR & \bl{$\mathbf{69.8}_{-2.2}^{+2.2}$} & 95.3 & 66.0 & \bf 63.4 & \bf 40.8 & \bf 32.2 \\
        \genie-large-6-6 &  \bl{$\mathbf{70.6}^{+2.1}_{-2.1}$} 	& \bf 95.1	&  \bf 66.3	& 63.1	& 40.7	& 26.3 \\
        \genie-base-6-6 & \bl{$65.0^{+2.3}_{-2.3}$}	& 94.7	& 64.9	& 61.3	& 38.6	& 16.8 \\ 
        JHU & \bl{$66.0^{+2.2}_{-2.2}$}	& 95.0	& 64.5	& 61.5	& 38.1	& 25.7 \\
        \bottomrule
    \end{tabular}
    \begin{tabular}{ccccccc}
        \multicolumn{7}{c}{ \emph{WMT21 (Machine Translation)} } \\
        \midrule
        Systems & Human & BERTScore & ROUGE & METEOR & SacreBLEU & BLEURT \\
        \midrule
        Watermelon & \bl{$\mathbf{75.7}_{-2.0}^{+2.0}$} & \bf 95.0 & \bf 64.8 & \bf 59.3 & \bf 34.5 & \bf 34.7 \\
        VolcTrans-AT  &  \bl{$\mathbf{75.2}^{+2.0}_{-2.0}$} 	& \bf 95.0	&  \bf 64.8	& \bf 59.3	& 34.4	& 34.6 \\
        HUMAN & \bl{$\mathbf{75.2}^{+2.0}_{-2.0}$}   & 94.8& 59.3 & 54.3&29.5&30.0\\
         \genie-large-6-6 & \bl{$70.4^{+1.9}_{-2.0}$}	& 94.9	& 63.3	& 57.0	& 32.4	& 31.3\\ 
         \genie-base-6-6 & \bl{$69.0^{+2.1}_{-2.1}$}	& 94.7	& 63.3	& 56.8	& 31.8 & 28.2 \\ 
        \genie-base-3-3 & \bl{$65.3^{+2.3}_{-2.3}$}	& 94.5	& 62.7	& 56.3 & 31.2	& 23.9 \\ 
        \genie-base-1-1 & \bl{$50.7^{+2.3}_{-2.4}$}	& 93.3	& 59.3	& 50.9	& 27.0	& -0.2 \\
    \end{tabular}
    \begin{tabular}{ccccccc}
        \toprule
        \multicolumn{7}{c}{ \emph{$\alpha$NLG (Commonsense reasoning)} } \\
        \midrule
        Systems & Human & BERTScore & ROUGE & METEOR & SacreBLEU & BLEURT \\
        \midrule
        T5 (11B) & \bl{$\mathbf{{75.9}}^{+1.1}_{-1.0}$} & \bf 92.9 & \bf 44.6 & \bf 35.2 & \bf 19.5 & \bf -22.2 \\
        GPT-2 (unsupervised) & \bl{$45.1^{+1.2}_{-1.3}$} & 88.5 & 19.7 & 18.8 & 1.8 & -84.5 \\
    \end{tabular}
    \resizebox{1\textwidth}{!}{
    \begin{tabular}{ccccccccccc}
        \toprule
        \multicolumn{11}{c}{ \large \emph{XSUM (Summarization)} } \\
        \midrule
        Systems & 
        \begin{tabular}{@{}c@{}}Human \\overall\end{tabular} & 
        \begin{tabular}{@{}c@{}}Human \\conciseness\end{tabular} &
        \begin{tabular}{@{}c@{}}Human \\fluency\end{tabular} & 
        \begin{tabular}{@{}c@{}}Human \\no-hallucination\end{tabular} & 
        \begin{tabular}{@{}c@{}}Human \\informativeness\end{tabular} & 
        BERTScore & ROUGE & METEOR & SacreBLEU & BLEURT \\
        \midrule
        Pegasus & \bl{$\mathbf{48.7}^{+3.1}_{-3.4}$} & 	\bl{$\mathbf{52.0}^{+2.3}_{-2.5}$}	& 	\bl{${49.1}^{+2.5}_{-2.4}$}	& 	\bl{$\mathbf{49.3}^{+2.9}_{-2.9}$}	& \bl{$\mathbf{49.2}^{+3.0}_{-2.8}$} & 91.9 & 39.1 & 35.4 & 16.7 & -17 \\
        T5 (11B) & \bl{$\mathbf{47.5}^{+3.3}_{-3.3}$} & \bl{${49.3}^{+2.1}_{-1.9}$} & \bl{$\mathbf{49.9}^{+2.7}_{-2.8}$} & \bl{$\mathbf{49.4}^{+2.8}_{-2.8}$} & \bl{${47.6}^{+3.0}_{-2.8}$} & \bf 92.0 & 37.9 & \bf 36.9 & 17.1 & \bf -14.3 \\ 
        \bottomrule
    \end{tabular}
    }
    \caption{Summary of evaluating several existing models on each dataset with \genie{}. The highest numbers (and their CI) in each column are indicated in {\bf bold}. The scores given by crowd workers are indicated with {\color{blue}blue} color. We evaluated all 24 systems from WMT21 but only show the top 3 systems as well as our \genie transformer baselines here.}
    \label{tab:main:experiments:appendix}
\end{table*}

\end{document}